%% file: acl_latex.tex
\documentclass[11pt]{article}

\usepackage[final]{acl}

\usepackage{times}
\usepackage{latexsym}

\usepackage[T1]{fontenc}

\usepackage[utf8]{inputenc}

\usepackage{microtype}

\usepackage{inconsolata}

\usepackage{graphicx}

\usepackage{amsmath}
\usepackage{amsfonts}
\usepackage{algpseudocode}
\usepackage{longtable}
\usepackage{makecell}
\usepackage{booktabs}
\usepackage{multirow}
\usepackage{dcolumn}
\usepackage{siunitx}
\usepackage{framed}
\usepackage{xcolor}
\usepackage{caption}
\usepackage[ruled,vlined]{algorithm2e}
\newcommand{\ours}{FedCoT}
\newcommand{\equalcontrib}{\textsuperscript{*}}
\newcommand{\corresponding}{\textsuperscript{†}}
\newcommand{\blfootnote}[1]{%
  \begingroup
  \renewcommand\thefootnote\relax
  \footnotetext{#1}%
  \endgroup
}
\title{FedCoT: Communication-Efficient Federated Reasoning Enhancement for Large Language Models}

\author{
    \textbf{Chuan Li\textsuperscript{1}\equalcontrib},
    \textbf{Qianyi Zhao\textsuperscript{1}\equalcontrib},
    \textbf{Fengran Mo\textsuperscript{2}},
    \textbf{Cen Chen\textsuperscript{1}\corresponding}  \\
    \textsuperscript{1}East China Normal University, Shanghai, China \\
    \textsuperscript{2}University of Montreal, Montreal, Canada \\
    \texttt{\{lichuan, 51255903037\}@stu.ecnu.edu.cn} \\
    \texttt{fengran.mo@umontreal.ca, cenchen@dase.ecnu.edu.cn}
}

\begin{document}
\maketitle
\blfootnote{* Equal contribution.}
\blfootnote{† Corresponding author.}

\input{main}

\end{document}

%% file: main.tex
\input{sections/abstract}

\input{sections/introduction}

\input{sections/related_works}
\input{sections/preliminaries}
\input{sections/methodology}
\input{sections/experiments}

\input{sections/conclusion}

\newpage

\input{sections/limitations}

\bibliography{custom}




\appendix

\clearpage

\input{sections/appendix/algorithm}

\input{sections/appendix/length}
\input{sections/appendix/explore_prm}
\input{sections/appendix/open_rm}

\input{sections/appendix/case_study}
\input{sections/appendix/model_improvement_potential}

\input{sections/appendix/detailed_datasets_info}

\input{sections/appendix/prompt_template}

\section{Sensitivity to Client Granularity}

The primary experiments use five cross-silo clients, with each biomedical QA dataset treated as one client. To examine a more fine-grained federation, we construct an eight-client setting by partitioning MedMCQA into two client shards and MedQA into three client shards, while retaining BioASQ, MMLU, and PubMedQA as individual clients. The resulting federation contains one BioASQ client, two MedMCQA clients, three MedQA clients, one MMLU client, and one PubMedQA client.

Table~\ref{tab:8client} compares the original five-client setting with the eight-client setting. The average accuracy changes from 55.50\% to 55.40\%, while the performance on individual benchmarks remains comparable. This result suggests that FedCoT remains stable under a moderately more fine-grained cross-silo partition. 

\begin{table*}[]
\centering
\begin{tabular}{lccccc c}
\toprule
\textbf{Method} & \textbf{BioASQ} & \textbf{MedMCQA} & \textbf{MedQA} & \textbf{MMLU} & \textbf{PubMedQA} & \textbf{Avg.} \\
\midrule
LlaMA-3-8B-Instruct & 37.90 & 29.80 & 27.20 & 38.70 & 9.20 & 28.56 \\
FedCoT (5 clients)  & 68.50 & 45.20 & 54.10 & 68.70 & 41.00 & 55.50 \\
FedCoT (8 clients)  & 69.40 & 44.30 & 52.80 & 69.30 & 41.20 & 55.40 \\
\bottomrule
\end{tabular}
\caption{Performance under different client granularities. The eight-client setting partitions MedMCQA into two clients and MedQA into three clients. All results are accuracy (\%).}
\label{tab:8client}
\end{table*}

\section{Effect of Shared Classifier-Head Aggregation}

FedCoT updates only the lightweight discriminator, including its LoRA modules and classification head, while keeping the actor LLM frozen. This design focuses communication on the component that directly evaluates and selects candidate reasoning paths. To examine the role of classifier-head sharing, we compare the default shared-head design with a local-head variant. The latter retains a separate classifier head on each client and aggregates only the discriminator LoRA modules.

Table~\ref{tab:head} reports the results with Qwen2.5-7B-Instruct as the actor model. Aggregating the classifier head improves the average accuracy from 52.40\% to 68.70\%. The shared-head variant also achieves higher accuracy on all five benchmarks. These results indicate that sharing the classifier head transfers cross-client calibration information that is complementary to the knowledge captured by the aggregated LoRA modules.

\begin{table*}[]
\centering
\begin{tabular}{lccccc c}
\toprule
\textbf{Method} & \textbf{BioASQ} & \textbf{MedMCQA} & \textbf{MedQA} & \textbf{MMLU} & \textbf{PubMedQA} & \textbf{Avg.} \\
\midrule
Qwen2.5-7B-Instruct          & 73.40 & 43.70 & 29.50 & 50.30 & 38.80 & 47.14 \\
FedCoT (shared head)         & 96.00 & 51.70 & 53.40 & 74.20 & 68.20 & 68.70 \\
FedCoT (personalized head)   & 80.60 & 48.20 & 29.50 & 60.10 & 43.60 & 52.40 \\
\bottomrule
\end{tabular}
\caption{Effect of classifier-head aggregation. The local-head variant keeps the classifier head client-specific and aggregates only LoRA modules. All results are accuracy (\%).}
\label{tab:head}
\end{table*}

\section{Analysis of Candidate Sampling Numbers}

\begin{figure}[h]
    \centering
    \includegraphics[width=1\linewidth]{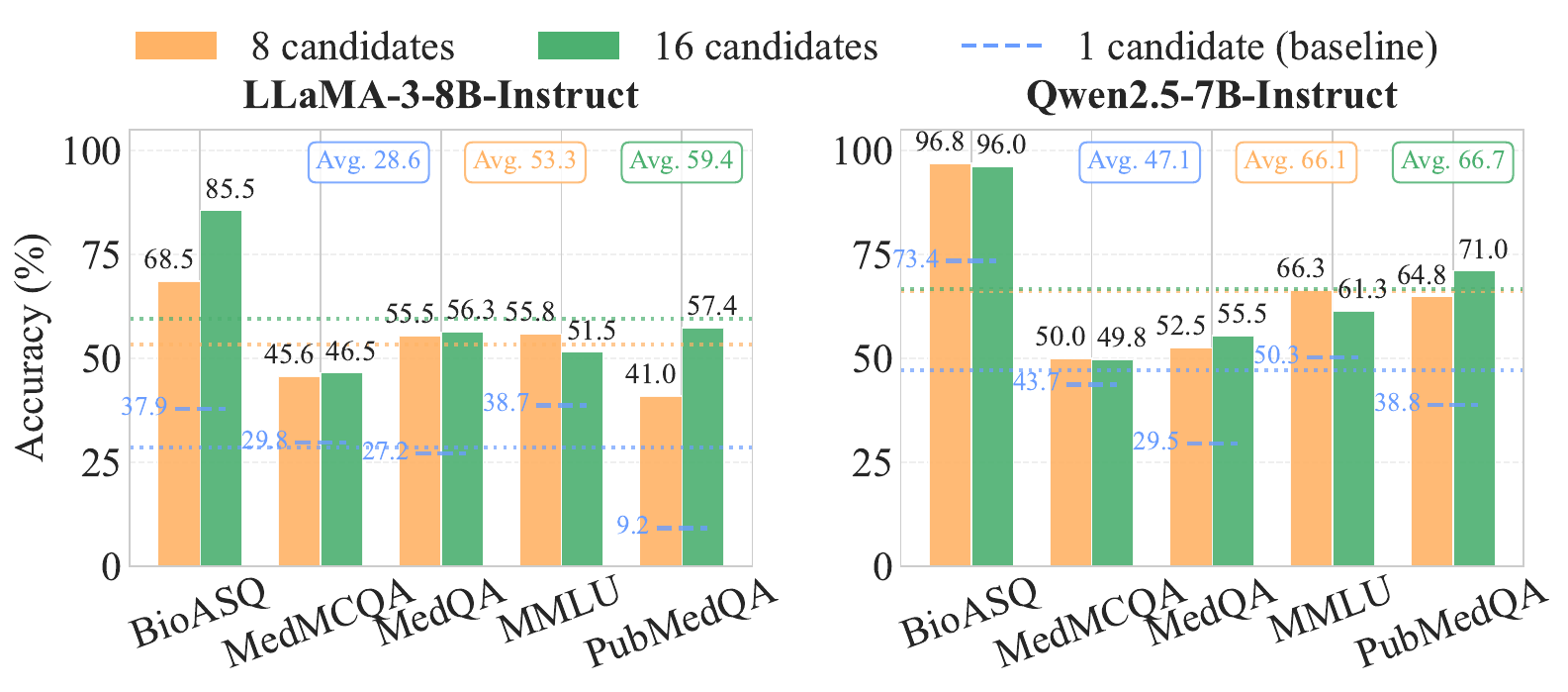}
    \caption{Performance improvement on difference candidate numbers of \ours{}.}
    \label{fig:exp_on_num}
\end{figure}

In candidates generation, the sampling number of candidates is sensitive to model performance. Our experimental analysis, illustrated in Figure~\ref{fig:exp_on_num}, demonstrates that increasing candidate samples from 8 to 16 consistently enhances performance across models. 
This improvement is particularly pronounced for \texttt{LLaMa-3-8B-Instruct}, where average accuracy increases from 52.32\% to 59.44\%. Notably, on the BioASQ dataset, its accuracy rises substantially from 65.30\% to 85.50\%. 

Although \texttt{Qwen2.5-7B-Instruct} also exhibits improvement gains (from 66.08\% to 66.72\%), the marginal improvement suggests model-dependent sensitivity to candidate quantity. In particular, the performance gap between 8 and 16 candidates is significantly smaller than that between 1 and 8 candidates for all backbone models, indicating that 8 candidates sufficiently demonstrate methodological efficacy.

\section{Assessing via LLM-as-Judge: Factuality, Explainability, Quality}

\begin{figure}[h]
    \centering
    \includegraphics[width=1\linewidth]{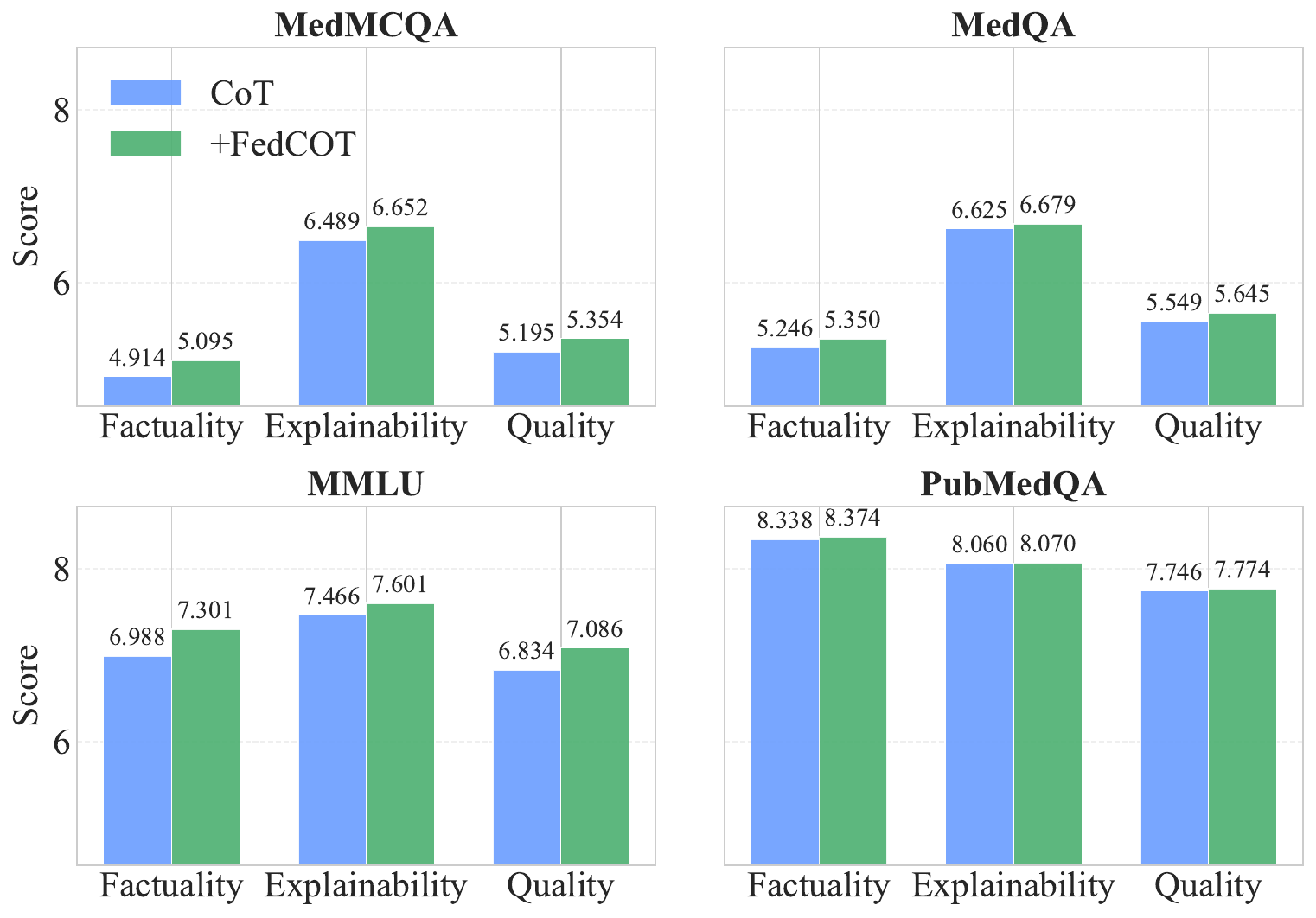}
    \caption{LLM-as-Judge evaluation on four biomedical QA datasets across three axes (higher is better): medical factuality, explainability/coherence, and overall quality.}
    \label{fig:llm_as_judge}
\end{figure}

We assess response quality using an LLM-as-Judge protocol with a strong judge model (\texttt{DeepSeek-V3.2-Exp}), which scores generated solutions along three axes-medical factuality, explainability/coherence, and overall quality-on a 1-10 scale; prompt details and rubric are in Appendix~\ref{ap:prompts}. Figure~\ref{fig:llm_as_judge} summarizes results on PubMedQA, MedMCQA, MedQA, and MMLU. \ours{} consistently improves over direct CoT across all datasets and axes. The largest improvements appear on MMLU (+0.31 factuality, +0.25 quality) and MedMCQA (+0.18 factuality, +0.16 explainability), while PubMedQA shows smaller but uniformly positive gains (+0.04 factuality, +0.01 explainability, +0.03 quality). These findings align with the accuracy improvements in Section~\ref{sec:main_results} and indicate that, despite using only final correctness as supervision during federated training, \ours{} effectively steers inference-time selection toward higher-quality reasoning chains.

\section{Performance on Smaller Size LLMs}

\begin{figure}[h]
    \centering
    \includegraphics[width=1\linewidth]{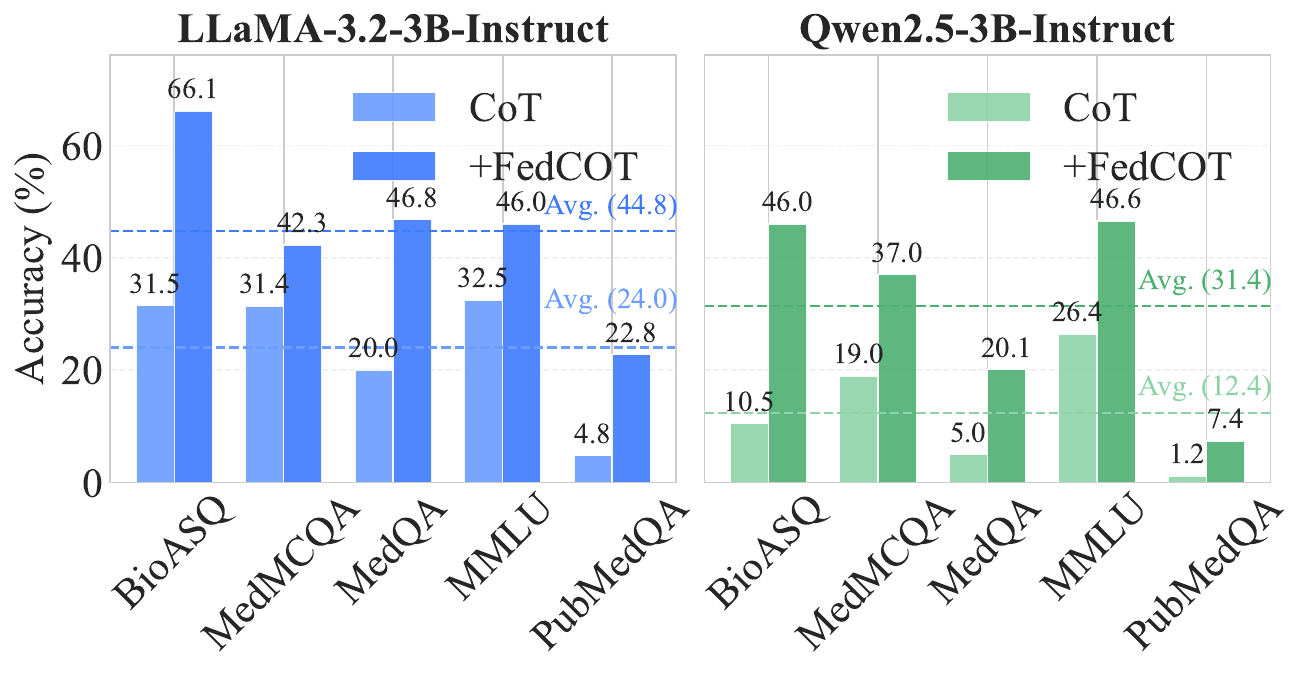}
    \caption{Performance improvement on 3B LLMs via federated reasoning fine-tuning on top of our \ours{}.}
    \label{fig:imp_on_slm}
\end{figure}

We further investigate the performance on smaller size LLMs, and the results are presented in Figure~\ref{fig:imp_on_slm}.
We can observe that our \ours{} exhibit moderate yet consistent gains on smaller models. Specifically, \ours{} significantly outperforms the CoT method with specific training on our federated learning framework and achieves an average performance improvement of 20.8\% on \texttt{LLaMA-3.2-3B-Instruct} and 19.00\% on \texttt{Qwen2.5-3B-Instruct}
These results underscore our methods' robust generalization and adaptability across a spectrum of model sizes.

\section{OOD Generalization}

\begin{figure}[h]
    \centering
    \includegraphics[width=1\linewidth]{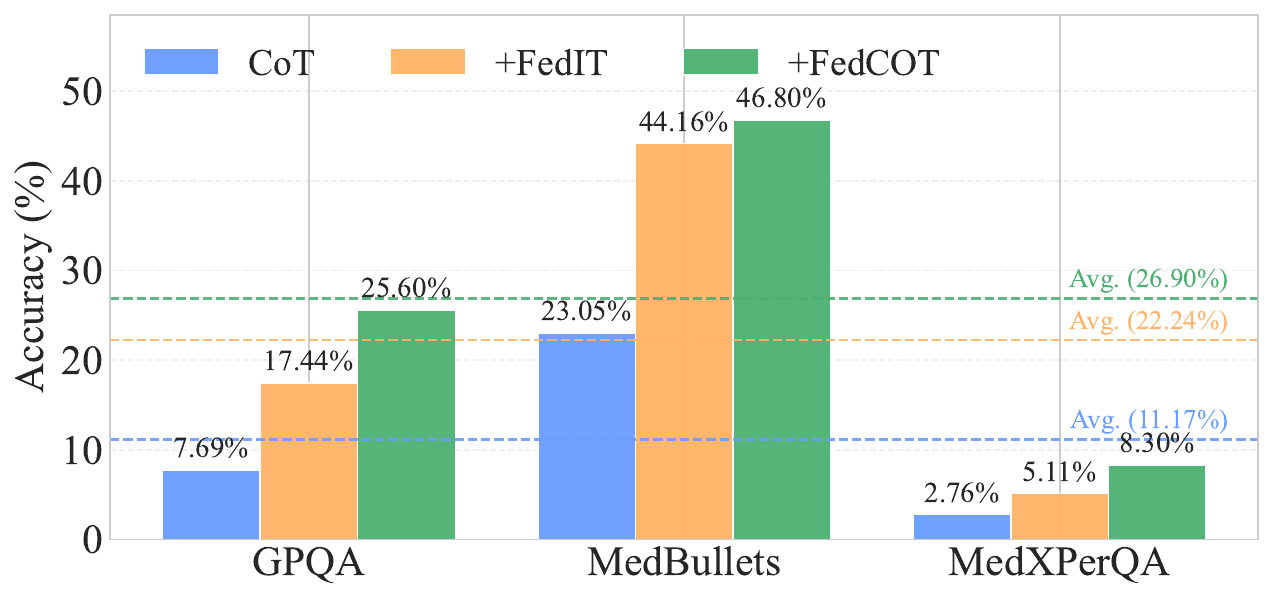}
    \caption{The validation in out-of-distribution situation.}
    \label{fig:ood}
\end{figure}

To assess generalization beyond the training distribution, we evaluate on three out-of-distribution (OOD) benchmarks-GPQA (biomedical track), MedBullets, and MedXPertQA. \ours{} maintains strong performance and consistently surpasses both the CoT and FedIT baseline on all benchmarks, as shown in Figure~\ref{fig:ood}. Specifically, \ours{} attains 25.60/46.80/8.30\% on GPQA/MedBullets/MedXPerQA, compared with FedIT’s 17.44/44.16/5.11\% and CoT’s 7.69/23.05/2.76\%. Averaged over the three datasets, \ours{} reaches 26.90\%, yielding a +4.66\% improvement over FedIT and a +15.73\% gain over CoT; the gains are most pronounced on GPQA (+8.16\% vs. FedIT). These results indicate that \ours{} offers better OOD robustness and transfers more effectively to other biomedical scenario under distribution shift.

%% file: sections/abstract.tex
\begin{abstract}

Enhancing LLM reasoning in federated settings is nontrivial due to stringent computational, communication, and privacy constraints, especially in healthcare, where clinically consequential decisions require not only accuracy but also interpretable, auditable rationales to meet safety, accountability, and regulatory requirements.
Conventional federated fine-tuning largely imitates final answers rather than cultivating step-by-step reasoning, often relying on privacy-sensitive centralized distillation and still incurring substantial communication overhead.
We address this gap with \textbf{\ours{}}, a federated reasoning framework that combines lightweight chain-of-thought resampling with a compact discriminator for selection, and client-aware LoRA stacking with weighted classifier aggregation to accommodate heterogeneity while reducing aggregation noise and communication; clients generate candidate chains and supervision locally, and only lightweight modules are aggregated on the server.
Experiments on medical reasoning benchmarks show consistent gains under tight resource budgets while keeping data local and respecting privacy, offering an interpretable and resource-efficient solution.
Our code is made publicly available at \url{https://github.com/DIaacKr/FedCoT}



\end{abstract}

%% file: sections/introduction.tex
\section{Introduction}


The development of large language models (LLMs) achieve advancing performance in complex reasoning tasks~\cite{touvron2023llama,bai2023qwen,guo2025deepseek,team2025kimi,chen2024huatuogpt}, which improves both the effectiveness and explainability based on the thought chains~\cite{wei2022chain}.
The promising performance is largely attributed to the reinforcement learning (RL) algorithms~\cite{christiano2017deep,schulman2017proximal,shao2024deepseekmath,zhou2025reinforced}.
However, the training paradigms of RL for reasoning models rely heavily on computational resources~\cite{tian2023learn, havrilla2024teaching}, which render them impractical for distributed edge environments, especially under the privacy constraint, e.g., the training data cannot be directly shared across different institutions in medical scenarios~\cite{chen2023customized}.

Training-free techniques~\cite{wang2022self, xie2023self, wu2024inference} with prompting engineering or test-time scaling can alleviate the issues of distributing the data and model during the training phase.
Although they are easy to deploy, their performance gains are quite limited and cannot fully leverage the collaborative potential of distributed device networks.
Intuitively, federated learning (FL)~\cite{mcmahan2017communication} serves as an alternative for achieving a better trade-off between privacy preservation guarantee and model performance. 
However, existing FL-based LLMs training paradigms~\cite{wu2025survey, wei2025federated, zhang2024towards} predominantly rely on federated supervised fine-tuning or simply incorporate parameter-efficient techniques, which still encounter high communication overhead and thus yield suboptimal performance gains.

Reasoning with Chain-of-Thought (CoT) is indispensable in medical domains. For example, the decision for clients should be not only accurate but also reliable with traceable rationales.
In addition, these privacy-sensitive scenarios demand more strict data and model usage to prevent sensitive information leakage.
However, existing studies~\cite{magister2022teaching,li2022explanations,wang2023scott,chen2024learning} usually obtain the data and rationales by distilling knowledge from proprietary models or direct sharing among different sources, which might expose sensitive information and violate data privacy principles. 

To facilitate reasoning-based LLMs training under a privacy-preserving setting without sharing the whole model in distributed nodes, we propose~\textbf{\ours{}}, a federated learning based framework to enhance the model's reasoning capacity under CoT prompting without data leakage.
The core of our \ours{} is a dynamic chain-of-thought discrimination mechanism to enable cross-client reasoning enhancement. 
Specifically, a lightweight discriminator is deployed to evaluate reasoning path candidates in real-time, which enables the discrimination of optimized reasoning trajectories.
Building upon the FLoRA algorithm's principal mechanism of modular LoRA stacking for federated fine-tuning~\cite{wang2024flora}, we incorporate a task-specific encoder-based discriminator~\cite{devlin2019bert} dedicated to reasoning path discrimination. To aggregate classifier parameters across clients, we employ a weighted aggregation scheme, thereby integrating discriminative knowledge under client heterogeneity.
By keeping raw data local and exchanging only lightweight discriminator updates, \ours{} supports data-local federated reasoning while maintaining robust performance under heterogeneous, privacy-sensitive settings. Experiments on five biomedical QA benchmarks show that \ours{} achieves the highest average accuracy among the evaluated methods for both backbone LLMs, while substantially reducing communication overhead relative to federated SFT baselines.

Our contributions are summarized as follows:
\begin{itemize}
\item To the best of our knowledge, we are the first to employ CoT resampling in federated learning settings to enhance the reasoning capabilities of large language models.
\item We propose an end-to-end federated reasoning enhancement framework that integrates dynamic reasoning path discrimination. \ours{} extends modular stacking to BERT-based discriminators with a weighted aggregation scheme, effectively handling client heterogeneity while maintaining robust performance.
\item We conduct comprehensive experiments across multiple medical QA benchmarks. The results demonstrate that \ours{} significantly enhances reasoning performance and efficiency compared to strong baselines, robustly validating the effectiveness of our federated reasoning enhancement framework.
\end{itemize}


%% file: sections/related_works.tex
\section{Related Works}

\paragraph{Reasoning Enhancement for Large Language Models}
CoT prompting, proposed by~\citet{wei2022chain}, acts as an effective mechanism to augment LLMs' reasoning ability with provided instructions, which
has spawned numerous training-free variants~\cite{wang2022self, chen2024self,li2025revisiting,nair2024midgard,wan2024reasoning,guo2025dsvd}. 
The follow-up studies integrate CoT with parameter updates by using CoT-generated rationales for model supervised fine-tuning ~\cite{kim2023cot,magister2022teaching,li2022explanations,wang2023scott,hsieh2023distilling}, or training with reinforcement learning~\cite{team2025kimi,shao2024deepseekmath}. 
However, the assumption of these methods is centralized data access, which neglects privacy constraints and computational burdens of generating rationales for federated clients~\cite{mcmahan2017communication}, and thus offering no tailored mechanism for CoT capability enhancement under distributed settings~\cite{zhang2024comprehensive,dritsas2025federated,wang2024feddse,li2019fair}.
To this end, our method utilizes the information of distributed clients to enhance the CoT reasoning ability while strictly ensuring privacy and low resource consumption.

\paragraph{Federated Learning for Large Language Models}
FL is one of the key solutions for LLMs training under privacy-preserving settings~\cite{wei2025federated,wu2025survey,chen2022learning,wu2020theoretical,tariq2023trustworthy,tariq2024trustworthy, ye2023robust,qian2024toward}.
The gradients and data aggregation during the training phase result in the communication-efficient needs and are usually achieved by parameter-efficient tuning.
For example, low-rank adaptation (LoRA) techniques \cite{zhang2024towards} and matrix factorization methods \cite{wang2024flora} enable efficient aggregation under heterogeneity. Besides, knowledge distillation \cite{fan2024fedmkt} and federated RL \cite{tian2023learn} provide other alternatives as optimization pathways. 
However, 
existing FL-LLM studies neither explicitly enhance CoT reasoning capabilities nor mitigate the privacy risks inherent in distilling rationales from centralized teacher models~\cite{havrilla2024teaching,li2022explanations}. In addition, federated RL-based approaches incur prohibitive computational and communication overhead for resource-constrained clients~\cite{qi2021federated,krouka2021communication,imteaj2022federated}. 
Thus, previous studies remain a critical gap for developing lightweight, privacy-preserving frameworks tailored to federated CoT enhancement, which our studies provide the exploration.

%% file: sections/preliminaries.tex
\section{Preliminaries}
\label{sec:preliminaries}

\subsection{Chain-of-Thought Prompting}
Chain-of-Thought (CoT) prompting~\cite{wei2022chain} enhances complex reasoning by generating intermediate reasoning paths \(\tau\) between input \( x \) and output \( y \) as guidance and explanation. Formally, given a model parameterized by \( \psi \) and prompt \(I\), the CoT generation process is defined as:  
\[ [\tau, y] \sim p_\psi(\tau, y | x,I) \]  
Although CoT is effective in a range of scenarios, utilizing it in distributed settings alone cannot perform well without leveraging collaborative devices for specific training.

\subsection{LoRA Federated Aggregation}
Low-Rank Adaptation (LoRA)~\cite{hu2022lora,mao2024dora} enables efficient fine-tuning via parameters metric decomposition \(\Delta W = BA\) (\(A \in \mathbb{R}^{r\times n}, B \in \mathbb{R}^{m\times r}\)). Standard federated learning conducts client model updates by averaging aggregation as 
\begin{equation}
\mathbf{A} = \sum_i^N u_i \mathbf{A}_i, \quad \mathbf{B} = \sum_i^N u_i \mathbf{B}_i
\end{equation}
where \(N\) denotes the number of clients, \(u_i\) denotes the weight of the client derived from the data volume ratio. 
This aggregation schema introduces cross-client noise terms as 
\begin{equation}
\begin{aligned}
\Delta\mathbf{W} & =(u_0\mathbf{B}_0+u_1\mathbf{B}_1)(u_0\mathbf{A}_0+u_1\mathbf{A}_1) \\
& = u_0^2\mathbf{B}_0\mathbf{A}_0+u_1^2\mathbf{B}_1\mathbf{A}_1 \\
&+\underbrace{u_0u_1(\mathbf{B}_0\mathbf{A}_1+\mathbf{B}_1\mathbf{A}_0)}_\text{noise term}
\end{aligned}
\end{equation}
However, the global updates would be deviated as the noise term grows quadratically with local clients. Besides, it would result in a dimension mismatch between heterogeneous ranks, i.e., \(r_1 \neq r_2\), leading to parameter updating with average fails.

%% file: sections/methodology.tex
\section{Methodology}
\label{sec:methodology}

\begin{figure*}[!th]
    \centering
    \includegraphics[width=0.95\linewidth]{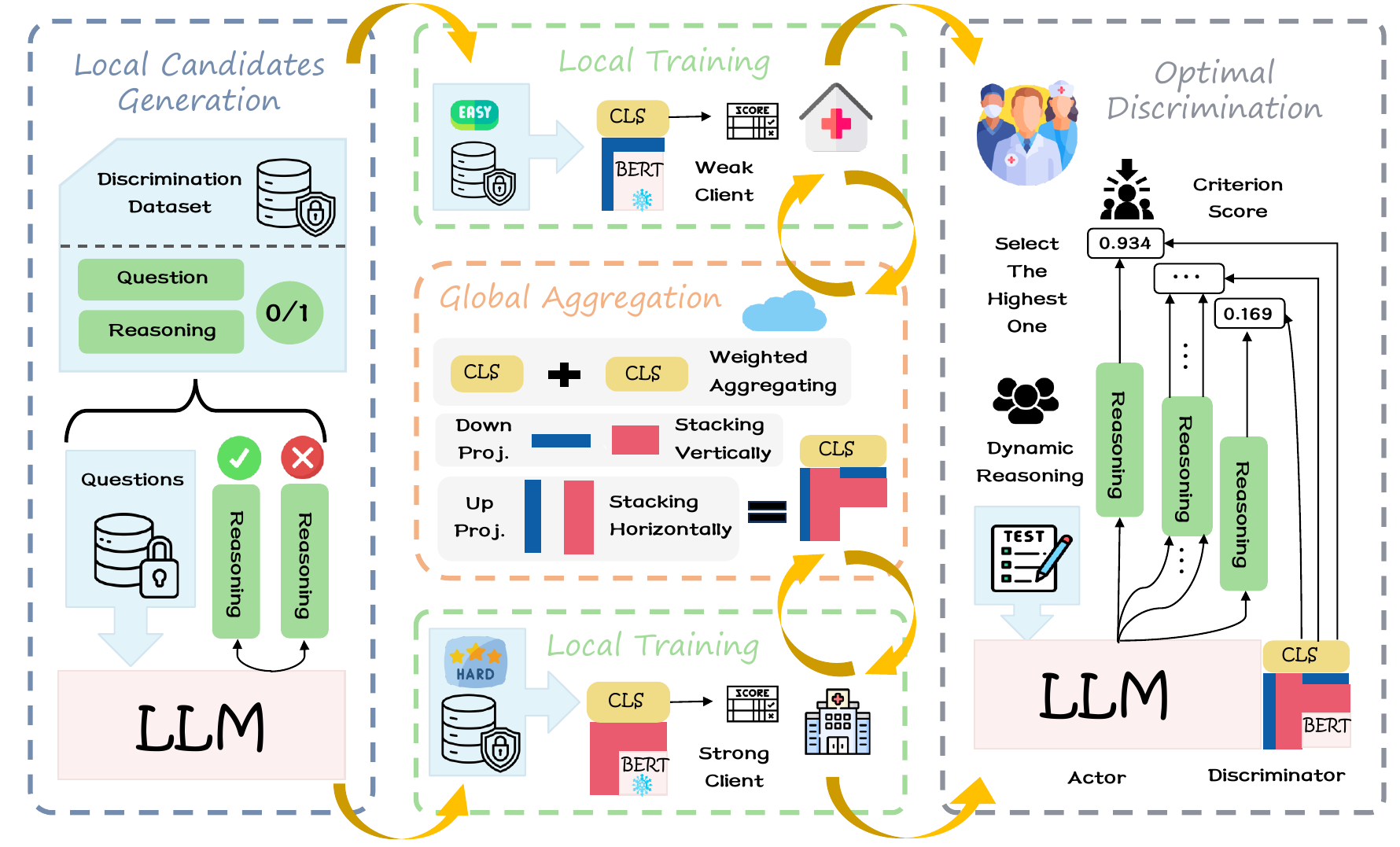}
    \caption{Overview of \ours{} framework. \textbf{Left:} The data preparation for training the discriminator. \textbf{Middle:} The federated fine-tuning of discriminator without the participation of local LLM which is only used in data preparation. \textbf{Right:} The Optimal Discrimination at test time. ``Down Proj.'' corresponds to the \(\mathbf A\) matrix in LoRA, and ``Up Proj.'' corresponds to the \(\mathbf B\) matrix. ``CLS" here denotes the classifier module of discriminator.
}
    \label{fig:overview}
\end{figure*}

We propose \ours{}, a federated learning-based framework to enhance the reasoning capability of LLMs under privacy constraints. 
The overview of \ours{} is presented in Figure~\ref{fig:overview} with dynamic path selection and parameter-efficient aggregation.
With the candidate reasoning paths generated locally by client LLM, the signals derived from these paths supervise the training of lightweight discriminators, whose LoRA modules and classifier are then aggregated within the server-side to construct a global discriminator. This federated model subsequently enables clients to dynamically select optimal reasoning traces during inference, yielding privacy-preserving, CoT-enhanced answers.


\subsection{Local Candidates Generation}
Under the federated learning framework, each client should first generate candidate reasoning paths based on the questions of their associated local datasets $\{x_j,y_j\}$, which then serve as the basis for the subsequent discriminator training.
Formally, $K$ candidate reasoning paths are generated by diversity sampling of an LLM $p_\psi$, which can also be analogously regarded as the actor model in a reinforcement-learning scenario, corresponding to the input $x_j$ as: 
\[
(\tau_{j,k}, \hat{y}_{j,k})
\sim p_\psi(\tau, \hat{y} \mid x_j, I).
\]

where each candidate answer $\hat{y}_{j,k}$ is assigned a binary correctness label by comparing it with the local ground-truth answer $y_j$, as defined in Eq.~\ref{eq: bin}. The resulting binary-labeled candidate set is constructed as Eq.~\ref{eq: d0}.
\begin{equation}
\label{eq: bin}
z_{j,k} = \mathbf{1}(\hat{y}_{j,k} = y_j), \quad k=1,2,...,K
\end{equation}
\begin{equation}
\label{eq: d0}
\mathcal{D}^{\mathrm{bin}} = \{ (\mathbf{h}_{j,k}, z_{j,k}) \mid \mathbf{h}_{j,k} = [x_j \parallel \mathbf{\tau}_{j,k} \parallel \hat{y}_{j,k}] \}
\end{equation}
To ensure high-quality supervision, we introduce a contrastive candidate selection mechanism to refine the dataset $\mathcal{D}^{\mathrm{bin}}$. We first discard queries with unanimously correct or incorrect paths, as they lack discriminative signals. For the remainder, we dynamically sample a balanced subset of positive and negative candidates, explicitly ensuring diverse error types (i.e., reasoning errors vs. missing answers) are represented.

\subsection{Local Training for Candidates Discrimination}
Following~\citet{shi2024medadapter}, we formulate reasoning-path discrimination as a binary classification task, 
where a lightweight encoder-based discriminator evaluates candidate correctness.

In our \ours{} framework, clients initialize their local models from one of the following choices: the server-provided global modules (for non-initial rounds) or a base pre-trained model locally (for the first round). 
During each global communication round, clients receive and initialize the model with the latest aggregated parameters, comprising both the LoRA matrices and classifier that encapsulate information from the entire federation while preventing the model drift from local domains.

Instead of relying on harsh binary labels like \cite{shi2024medadapter}, we design a soft label to evaluate the logical consistency between the reasoning path $\tau_{j,k}$ and the predicted answer $\hat{y}_{j,k}$. We compute the entailment probability $p^{\mathrm{entail}}_{j,k} \in [0,1]$ as NLI task (Natural Language Inference). The continuous soft label $s_{j,k}$ is formulated as a combination of the outcome correctness and the reasoning validity:
\begin{equation}
\label{eq: s_entail}
s_{j,k} = (1 - \alpha) \cdot \mathbf{1}(\hat{y}_{j,k} = y_j) + \alpha \cdot p^{\mathrm{entail}}_{j,k}
\end{equation}
where $\alpha=0.5$ is a balancing coefficient. This continuous label assigns partial credit to flawed answers with rigorous reasoning, while penalizing correct guesses lacking logical support. 

Given a candidate feature $\mathbf{h}_{j,k}$, the discriminator produces a probability score
\[
d_\phi(\mathbf{h}_{j,k})
=
\sigma\!\left(f_\phi(\mathbf{h}_{j,k})\right)
\in (0,1),
\]
where $f_\phi(\cdot)$ denotes the discriminator logit.
The discriminator is optimized using the binary cross-entropy loss with the soft target $s_{j,k}$:
\begin{equation}
\begin{aligned}
\mathcal{L} = 
& - \left[ s_{j,k} \log d_\phi(\mathbf h_{j,k}) + \right. \\
&\left. (1-s_{j,k}) \log (1 - d_\phi(\mathbf h_{j,k})) \right]
\end{aligned}
\label{eq:seq_cls_loss}
\end{equation}

The discrimination dataset is constructed as:
\begin{equation}
\label{eq: dis}
\mathcal{D} = \{ (\mathbf{h}_{j,k}, s_{j,k}) \mid \mathbf{h}_{j,k} = [x_j \parallel \mathbf{\tau}_{j,k} \parallel \hat{y}_{j,k}] \}
\end{equation}

\subsection{Modular Global Aggregation}
We adopt and integrate FLoRA~\cite{wang2024flora} to achieve noise-free aggregation of LoRA matrix with protecting data privacy.
When aggregating local LoRA modules, the global model update $\Delta W$ can be expressed as
\begin{equation}
\begin{aligned}
\Delta\mathbf{W} 
&=\sum_{i=1}^N\mathbf{B}_i\mathbf{A}_i \\
&=\left(\mathbf{B}_1\oplus\mathbf{B}_2\oplus\cdots\oplus\mathbf{B}_N\right)\\
&\quad\cdot \left(\mathbf{A}_1\oplus\mathbf{A}_2\oplus\cdots\oplus\mathbf{A}_N\right)
\end{aligned}
\label{eq:lora_stack}
\end{equation}
where ``$\oplus$'' represents the matrix stacking operation, i.e., stacking them vertically along the row direction for $\mathbf A_i$ and stacking horizontally along the column direction for $\mathbf B_i$, respectively.
With the principle of block matrix multiplication, the product of these two global produced matrices, $\mathbf{B} \cdot \mathbf{A}$, is mathematically equivalent to the sum of the individual local updates, $\sum_{i=1}^N \mathbf{B}_i\mathbf{A}_i$. 

This approach makes the globally aggregated discriminator more reliable and adaptable to heterogeneity, which arises from varying client capabilities (e.g., weaker clients using smaller LoRA ranks, stronger ones using larger ranks). We can also intentionally create heterogeneity by assigning smaller ranks to simpler tasks and larger ranks to complex ones. Regardless of the source, the stacking method integrates these diverse LoRA matrices through unified merging, ensuring smooth federated learning.

Besides, the classifier weights of each client are aggregated using a weighted average approach at each global round, to integrate information across downstream tasks as
\begin{equation}
\begin{aligned}
\mathbf{W}^{cls} = \sum_{i}^N u_i \mathbf{W}_i^{cls}
\end{aligned}
\label{eq:aggregate_cls}
\end{equation}

\subsection{Optimal Discrimination}
During the inference stage, each client utilizes the final global discriminator model to score the multiple candidate reasoning paths and then selects the one with the highest score as the final output to achieve dynamic reasoning as:
\begin{equation}
r(\mathbf{h}_{j,k}) = d_\phi(\mathbf{h}_{j,k}).
\label{eq:get_p}
\end{equation}

\begin{equation}
k_j^{*}
=
\arg\max_{k\in\{1,\ldots,K\}}
r(\mathbf{h}_{j,k}),
\label{eq:get_score}
\end{equation}
\begin{equation}
\hat{y}_j
=
\hat{y}_{j,k_j^{*}}.
\label{eq:get_answer}
\end{equation}

The comprehensive process is described as an Algorithm provided in the Appendix~\ref{ap:algo} as the overall process of federated reasoning in our \ours{}.

%% file: sections/experiments.tex
\section{Experiments}
\subsection{Experimental Setup}

\paragraph{Datasets}
We evaluate our method on five biomedical Question-Answering (QA) datasets served as privacy-preserving benchmarks following previous studies~\cite{chen2024huatuogpt,song2025optimizing,chen2024policy,zhang2023enhancing}, including BioASQ
\cite{tsatsaronis2015overview}, MedMCQA
\cite{pal2022medmcqa}, MedQA
\cite{jin2021disease}, MMLU-MED
\cite{hendrycks2020measuring}, and PubMedQA
\cite{jin2019pubmedqa}. 
The statistics are provided in Table~\ref{tab:datasets}.
These datasets span diverse task categories, ranging from medical examination questions to literature-based QA, enabling us to evaluate the models' performance across complex reasoning tasks comprehensively.
The detailed information is provided in Appendix~\ref{ap:detailed_datasets_info}.

\input{tables/datasets_info}

\paragraph{Cross-silos Setting} 
The five datasets are treated as independent clients, and raw client data remain local throughout training. This cross-silo setting reflects the model's reasoning capability on different tasks and verifies its robustness under data distribution heterogeneity, which aligns with the data islands~\cite{huang2021personalized,huang2023promoting,tang2021incentive,liu2022privacy} in the real-world setting.

\paragraph{Prompt Templates}
Our CoT template follows a standardized and structured format with explicit itemized requirements. 
We also incorporate a one-shot CoT demonstration in the template, where the completed information is provided in Appendix~\ref{ap:prompts}.

\input{tables/main_results}

\paragraph{Hyperparameter Settings}
We generate 8 candidate responses for each query sample with a maximum length of 512 tokens.
The LoRA ranks for each client model in BioASQ, MedMCQA, MedQA, MMLU, and PubMedQA datasets are set as 4, 32, 32, 16, and 4, respectively.
During supervised fine-tuning of the federated LLMs, a uniform LoRA with a rank of 32 was used for model training.
The global round is set to 2, and the local training epoch is set to 1 with a batch size of 2 in SFT of LLMs as baselines.
The global round is set to 3, and the local training epoch is set to 1 with a batch size of 16 in our discriminator training.

\paragraph{Baselines and Evaluation Metrics}
Our experiments utilize different models for evaluation, including \texttt{Qwen2.5-7B-Instruct}~\cite{bai2023qwen}, \texttt{LLaMA-3-8B-Instruct}~\cite{touvron2023llama}, as core LLMs for main evaluation, and the \texttt{Longformer-base-4096}~\cite{beltagy2020longformer} as discriminator model, following \citet{shi2024medadapter}. 
We compare with both the training-free and training-based baselines under both federated and non-federated scenarios to evaluate our \textbf{\ours{}} as follows:
We evaluate \textbf{\ours{}} against a diverse set of strong baselines, including \textbf{Self-Consistency}~\cite{wang2022self} as a training-free inference strategy via diverse sampling and majority voting. For federated learning baselines, we compare with \textbf{FedIT}~\cite{zhang2024towards} utilizing weighted averaging, \textbf{FedFFA-LoRA}~\cite{sun2024improving} for parameter-efficient tuning, \textbf{FLORA}~\cite{wang2024flora} designed for heterogeneous LoRA rank adaptations, and \textbf{FedPrompt}~\cite{zhao2023fedprompt} for communication-efficient prompt tuning.
Accuracy is adopted as the primary evaluation metric to be consistent with previous studies~\cite{chen2024huatuogpt}. 
All evaluations measure accuracy under CoT prompting, which not only quantifies performance but also aligns with real-world medical demands for interpretability and response safety via inherent step-by-step reasoning.

\subsection{Main Results}
\label{sec:main_results}

\begin{figure}[t]
    \centering
    \includegraphics[width=1\linewidth]{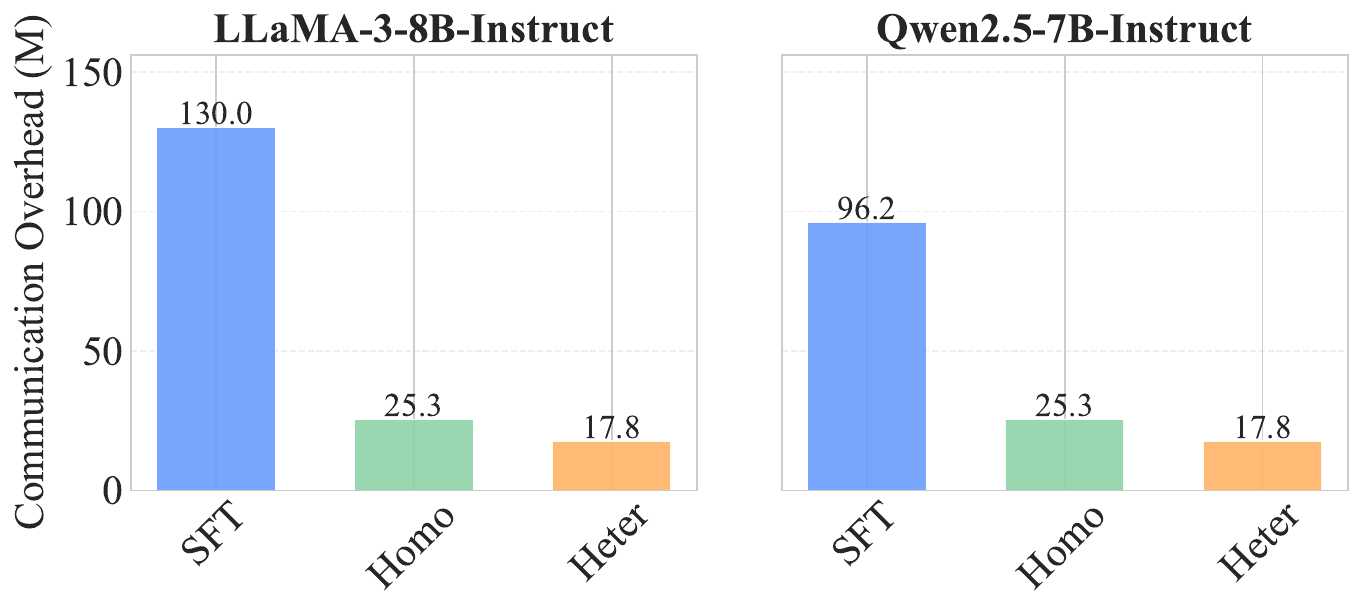}
    \caption{Analysis of communication efficiency in federated SFT and our \ours{}. ``SFT" represents FedIT, ``Homo" represents \ours{} with lora rank of 32, ``Heter" represents \ours{} with lora rank of 4, 32, 32, 16, 4.}
    \label{fig:commu}
\end{figure}

\paragraph{Overall Performance} 

The overall results are presented in Table \ref{tab:main_results}. \ours{} achieves the highest average accuracy among the evaluated methods on both backbone LLMs, demonstrating the overall effectiveness of our framework across diverse biomedical QA tasks.

Specifically, \ours{} achieves substantial average absolute improvements of 26.94\% and 21.56\% over the base \texttt{LLaMA-3-8B-Instruct} and \texttt{Qwen2.5-7B-Instruct} models with CoT prompting. Furthermore, our framework achieves the strongest average performance among the evaluated federated baselines by surpassing the strongest advanced federated fine-tuning baselines, outperforming FedPrompt by 9.40\% on LLaMA-3 and FLORA by 6.29\% on Qwen2.5. 

While other parameter-efficient federated methods (e.g., FedIT, FedFFA-LoRA, and FLORA) yield moderate gains over the base models, they struggle to generalize evenly across diverse medical reasoning tasks. In contrast, \ours{} demonstrates robust improvements across highly varied domains, from specialized literature (BioASQ, PubMedQA) to comprehensive clinical examinations (MedQA, MedMCQA). These results suggest that dynamic reasoning-path discrimination provides a strong alternative to generic federated fine-tuning under the evaluated cross-silo setting.

\paragraph{Efficiency Comparison}

\begin{table*}[htbp]
\centering
\resizebox{\textwidth}{!}{
\begin{tabular}{lcccccc}
\toprule
\textbf{Method} & \textbf{BioASQ} & \textbf{MedMCQA} & \textbf{MedQA} & \textbf{MMLU} & \textbf{PubMedQA} & \textbf{Avg.} \\
\midrule
Best-of-N (Logits-based)       & 40.32 & 27.76 & 26.24 & 38.04 & 5.60  & 27.59 \\
Best-of-N (Qwen3-0.6B-Reranker)& 49.20 & 26.50 & 29.60 & 36.20 & 12.40 & 30.78 \\
FedCoT (Local)                     & 57.30 & 42.60 & 40.80 & 56.40 & 29.20 & 45.26 \\
FedCoT (Qwen3-0.6B-Base)       & 50.00 & 43.70 & 37.00 & 49.70 & 13.20 & 38.72 \\
\midrule
FedCoT (ours)                      & 68.50 & 45.20 & 54.10 & 68.70 & 41.00 & 55.50 \\
\bottomrule
\end{tabular}}
\caption{Performance of different filtering method on 8 candidates. FedCoT (ours) denotes the full federated framework. FedCoT (Local) represents models trained only on individual client datasets. FedCoT (Qwen3-0.6B-Base) replaces the default discriminator with a generative backbone. All results are accuracy (\%).}
\label{tab:filtering-effects}
\end{table*}

\input{tables/federated_results}


The efficiency comparison is shown in Figure~\ref{fig:commu}, which compares the communication efficiency between federated SFT methods (FedIT) and our \ours{}. 
During the federated learning process, the Federated SFT method performs LoRA fine-tuning on the LLM actor \texttt{LLaMA-3-8B-Instruct} and \texttt{Qwen2.5-7B-Instruct}. Here, the values 130M and 96M refer to the total number of parameters that need to be transmitted across all clients in the federated system during one global round of training. However, even with LoRA fine-tuning, such a volume of parameters still incurs large computational and communication overheads for low-resource clients and the whole federated learning system. 
In contrast, our \ours{} greatly reduces the training and communication overheads by fine-tuning a lightweight model. 
Thus, the parameter quantity of \ours{} during the federated learning process only accounts for 25.3M and 17.8M, which is much more efficient than the compared existing SFT methods, demonstrating its efficiency in low-resource environments.
\subsection{Ablation Study}

\paragraph{Candidate Processing}
To evaluate our data processing components, we conducted ablation studies by independently removing Candidate Selection and Soft Labeling (Table \ref{tab:ablation}). Results show that omitting Selection introduces uninformative, imbalanced samples that degrade the discriminator’s decision boundaries. Similarly, replacing Soft Labeling with binary labels fails to distinguish between rigorous reasoning and ``lucky guesses.'' The full \ours{} framework achieves the highest average accuracy, particularly on the MMLU dataset, confirming that refined, smoothed supervision is essential for complex multi-disciplinary reasoning.



\begin{table*}[th!]
\centering
\resizebox{\textwidth}{!}{
\begin{tabular}{@{}l cc cc cc cc cc c@{}}
\toprule
\multirow{2}{*}{\textbf{Method}} 
& \multicolumn{2}{c}{\textbf{BioASQ}} 
& \multicolumn{2}{c}{\textbf{MedMCQA}} 
& \multicolumn{2}{c}{\textbf{MedQA}}
& \multicolumn{2}{c}{\textbf{MMLU}}
& \multicolumn{2}{c}{\textbf{PubMedQA}}
& \textbf{Avg.} \\
\cmidrule(lr){2-3} \cmidrule(lr){4-5} \cmidrule(lr){6-7} 
\cmidrule(lr){8-9} \cmidrule(lr){10-11} 
& Acc. (\%) & Size 
& Acc. (\%) & Size 
& Acc. (\%) & Size 
& Acc. (\%) & Size 
& Acc. (\%) & Size 
& Acc. (\%) \\
\midrule
\textbf{FedCoT (Full)}          & 68.50 & 364 & 45.20 & 1420 & 54.10 & 7278 & 68.70 & 915 & 41.00 & 205 & 55.50 \\
\quad w/o Candidate Selection & 69.40 & 494 & 44.90 & 2000 & 53.70 & 10178 & 62.90 & 1299 & 41.00 & 450 & 54.38 \\
\quad w/o Soft Labeling           & 67.70 & 364 & 45.10 & 1420 & 54.50 & 7278 & 60.30 & 915 & 41.00 & 205 & 53.72 \\
\quad w/o CS \& SL & 65.30 & 494 & 45.20 & 2000 & 56.10 & 10178 & 54.00 & 1299 & 41.00 & 450 & 52.32 \\
\bottomrule
\end{tabular}
}
\caption{Ablation study of \ours{} components. ``Size'' denotes the number of candidate paths retained for local discriminator training.}
\label{tab:ablation}
\end{table*}

\paragraph{Selection Strategies}
Table \ref{tab:filtering-effects} compares the filtering performance of our proposed FedCoT against various reward modeling strategies. FedCoT consistently achieves the highest accuracy (55.50\% on average), outperforming both the vanilla Logits-based baseline and specialized reranking models like Qwen3-0.6B-Reranker. Notably, our encoder-based federated approach surpasses its local counterpart, demonstrating the necessity of cross-client knowledge aggregation. Furthermore, although larger generative backbones (e.g., Qwen3-0.6B) or SOTA open-source reward models (detailed in Appendix \ref{ap:open_rm}) were tested, they failed to match our performance. This highlights that in low-resource scenarios, evaluating the complex deep semantics of small models may be less effective than assessing their relatively shallow semantics.

\subsection{Analysis of Heterogeneous LoRA Configurations}

\begin{figure}[h]
    \centering
    \includegraphics[width=0.95\linewidth]{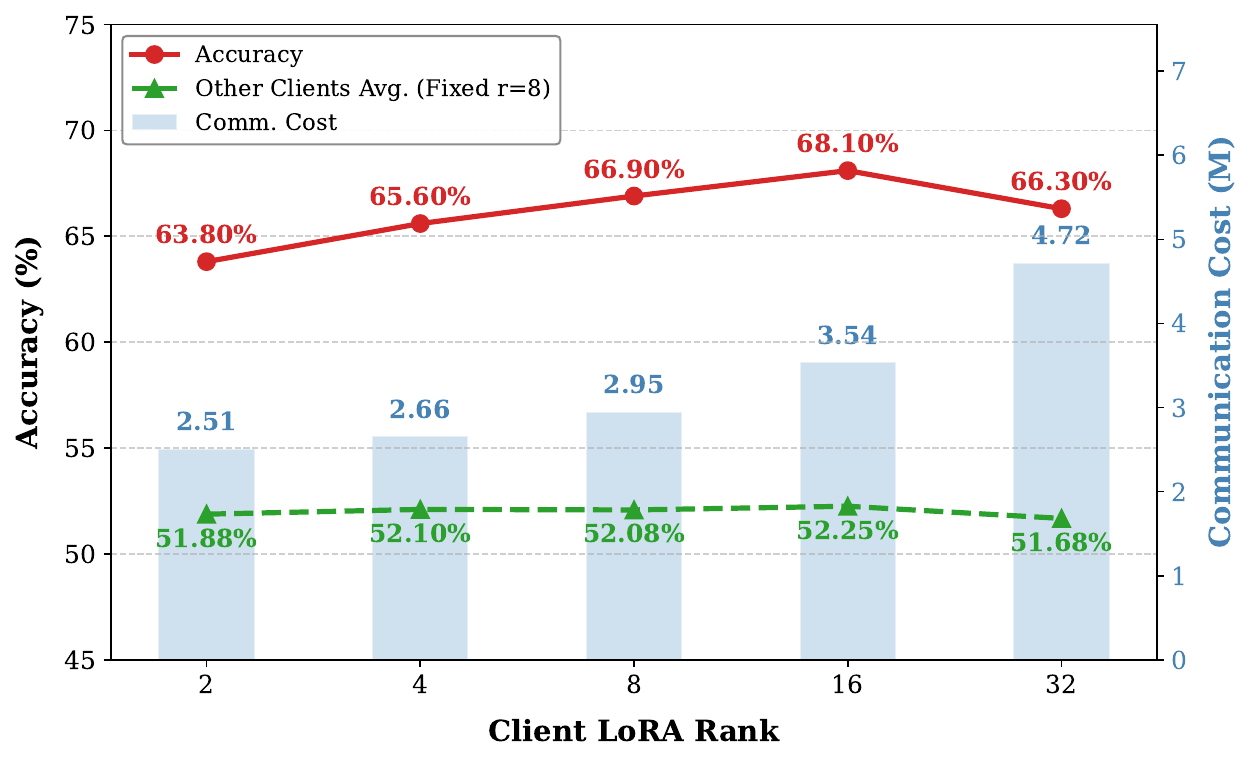}
    \caption{Impact of varying the MMLU client's LoRA rank on local accuracy, communication cost, and the average performance of other fixed-rank clients ($r=8$).}
    \label{fig:all8but}
\end{figure}

In real-world medical federated networks, clients often face extreme heterogeneity not only in local computational budgets and network bandwidths but also in the intrinsic complexity of their respective clinical tasks. Forcing a homogeneous LoRA rank across all clients is suboptimal, which wastes communication bandwidth and risks overfitting on simpler tasks, while creating severe capacity bottlenecks for complex medical reasoning tasks. 

To evaluate performance under heterogeneous rank configurations, we conduct a controlled stress test (Figure \ref{fig:all8but}). We fix the LoRA rank of four clients at $r=8$ while drastically varying the rank of the MMLU client from $r=2$ to $r=32$. As the results illustrate, reducing the MMLU rank from 16 down to 2 substantiallycuts its local communication cost from 3.54M to 2.51M parameters per round. Despite this aggressive downscaling, the client suffers only a smooth, acceptable degradation in accuracy from 68.10\% to 63.80\%. More importantly, throughout this compression phase, the average performance of the other four clients remains remarkably stable at approximately 52.10\%. This stability empirically proves that the adopted modular stacking mechanism effectively isolates local capability bottlenecks, preventing severe low-rank truncation noise from poisoning the global aggregated knowledge. Conversely, naively amplifying the MMLU rank to 32 maximizes the communication burden (4.72M) but triggers a counterintuitive overfitting penalty. Rather than improving reasoning capabilities, this over-parameterization not only degrades MMLU's local accuracy to 66.30\% but also inadvertently drags down the average performance of the remaining network to 51.68\%.

Guided by these systemic insights, a practical federated framework must flexibly accommodate heterogeneous ranks to adapt to diverse real-world conditions, such as varying computational budgets and task complexities. Enforcing a uniform rank across all clients inevitably forces a rigid compromise between communication efficiency and local model capacity. As shown in Table \ref{tab:federated_results}, by allowing clients to independently operate with varied ranks (e.g., $r \in \{4, 16, 32\}$) tailored to their specific local constraints, FedCoT integrates these asymmetric updates. Remarkably, this flexible heterogeneous configuration eliminates the unrealistic need for strict system-wide uniformity. This demonstrates that FedCoT elegantly adapts to diverse and constrained client environments while maximizing global reasoning stability.

%% file: tables/datasets_info.tex
\begin{table}[t]
    \centering
    \footnotesize  
    \begin{tabular}{lccc}
        \hline
        \textbf{Datasets} & \textbf{Train} & \textbf{Test} & \textbf{Source} \\
        \hline
        PubMedQA \citeyearpar{jin2019pubmedqa}            & 450   & 500   & Experts \\
        BioASQ \citeyearpar{tsatsaronis2015overview}      & 494   & 124   & Articles \\
        MMLU \citeyearpar{hendrycks2020measuring}         & 1299  & 163   & Examination \\
        MedMCQA \citeyearpar{pal2022medmcqa}              & 3000  & 4183  & Examination \\
        MedQA \citeyearpar{jin2021disease}                & 10178 & 1273  & Examination \\
        \hline
    \end{tabular}
    \caption{The size and source of the medical QA datasets used in the experiment.}
    \label{tab:datasets}
\end{table}

%% file: tables/main_results.tex
\begin{table*}[th!]
\centering
\resizebox{\textwidth}{!}{  %
\fontsize{8}{9.5}\selectfont  %
\setlength{\tabcolsep}{3pt}  %
\begin{tabular}{@{}l *{6}{cc} @{}}  %
\toprule
\multirow{2.5}{*}{\textbf{Method}} 
& \multicolumn{2}{c}{\textbf{BioASQ}} 
& \multicolumn{2}{c}{\textbf{MedMCQA}} 
& \multicolumn{2}{c}{\textbf{MedQA}}
& \multicolumn{2}{c}{\textbf{MMLU}}
& \multicolumn{2}{c}{\textbf{PubMedQA}}
& \multicolumn{2}{c}{\textbf{Avg.}} \\
\cmidrule(lr){2-3} \cmidrule(lr){4-5} \cmidrule(lr){6-7} 
\cmidrule(lr){8-9} \cmidrule(lr){10-11} \cmidrule(lr){12-13}
& \makecell{Acc. (\%)} & $\Delta$(\%) 
& \makecell{Acc. (\%)} & $\Delta$(\%) 
& \makecell{Acc. (\%)} & $\Delta$(\%) 
& \makecell{Acc. (\%)} & $\Delta$(\%) 
& \makecell{Acc. (\%)} & $\Delta$(\%) 
& \makecell{Acc. (\%)} & $\Delta$(\%) \\
\midrule

LLaMA-3-8B-Instruct & 37.90 & \textemdash & 29.80 & \textemdash & 27.20 & \textemdash & 38.70 & \textemdash & 9.20 & \textemdash & 28.56 & \textemdash \\
\addlinespace[0.2em]
\quad +Self-Consistency & 40.30 & +2.40 & 31.50 & +1.70 & 24.70 & -2.50 & 41.10 & +2.40 & 2.80 & -6.40 & 28.08 & -0.48 \\
\quad +FedIT          & 42.74 & +4.84 & \textbf{47.29} & \textbf{+17.49} & \underline{53.73} & \underline{+26.53} & \textbf{71.17} & \textbf{+32.47} & 13.20 & +4.00 & 45.63 & +17.07 \\
\quad +FedFFA-LoRA    & 25.00 & -12.90 & 39.47 & +9.67 & 37.78 & +10.58 & 57.67 & +18.97 & 4.60 & -4.60 & 32.90 & +4.34 \\
\quad +FLORA          & 37.10 & -0.80 & 42.05 & +12.25 & 45.56 & +18.36 & 61.96 & +23.26 & 7.00 & -2.20 & 38.73 & +10.17 \\
\quad +FedPrompt      & \underline{52.42} & \underline{+14.52} & \underline{45.80} & \underline{+16.00} & 49.96 & +22.76 & 42.94 & +4.24 & \underline{39.40} & \underline{+30.20} & \underline{46.10} & \underline{+17.54} \\
\quad +FedCoT (Ours)  & \textbf{68.50} & \textbf{+30.60} & 45.20 & +15.40 & \textbf{54.10} & \textbf{+26.90} & \underline{68.70} & \underline{+30.00} & \textbf{41.00} & \textbf{+31.80} & \textbf{55.50} & \textbf{+26.94} \\
\addlinespace[0.5em]
\midrule
Qwen2.5-7B-Instruct & 73.40 & \textemdash & 43.70 & \textemdash & 29.50 & \textemdash & 50.30 & \textemdash & 38.80 & \textemdash & 47.14 & \textemdash \\
\addlinespace[0.2em]
\quad +Self-Consistency & \underline{86.30} & \underline{+12.90} & 47.10 & +3.40 & 28.00 & -1.50 & 57.10 & +6.80 & 39.80 & +1.00 & 51.66 & +4.52 \\
\quad +FedIT          & 82.26 & +8.86 & 48.48 & +4.78 & 44.30 & +14.80 & 68.71 & +18.41 & 47.20 & +8.40 & 58.19 & +11.05 \\
\quad +FedFFA-LoRA    & 85.48 & +12.08 & 48.74 & +5.04 & 44.93 & +15.43 & 70.55 & +20.25 & 53.00 & +14.20 & 60.54 & +13.40 \\
\quad +FLORA          & 83.87 & +10.47 & \underline{51.30} & \underline{+7.60} & \underline{45.88} & \underline{+16.38} & \underline{73.62} & \underline{+23.32} & \underline{57.40} & \underline{+18.60} & \underline{62.41} & \underline{+15.27} \\
\quad +FedPrompt      & 66.13 & -7.27 & 46.62 & +2.92 & 44.23 & +14.73 & 68.71 & +18.41 & 50.40 & +11.60 & 55.22 & +8.08 \\
\quad +FedCoT (Ours)  & \textbf{96.00} & \textbf{+22.60} & \textbf{51.70} & \textbf{+8.00} & \textbf{53.40} & \textbf{+23.90} & \textbf{74.20} & \textbf{+23.90} & \textbf{68.20} & \textbf{+29.40} & \textbf{68.70} & \textbf{+21.56} \\
\addlinespace[0.5em]
\bottomrule
\end{tabular}%
}
\caption{Performance of different methods across five privacy-preserving medical datasets on top of two backbone LLMs under different settings. The best results are in \textbf{Bold} and the second-highest results are indicated with an \underline{underline}.}
\label{tab:main_results}
\end{table*}

%% file: tables/federated_results.tex
\begin{table*}[th!]
\centering
\begin{tabular}{@{}lcccccc@{}}
\toprule
Method & BioASQ & MedMCQA & MedQA & MMLU & PubMedQA & Avg.(\%) \\
\midrule
CoT & 37.90 & 29.80 & 27.20 & 38.70 & 9.20 & 28.56 \\
\midrule
FedCoT (r=4,4,4,4,4)  & 67.70 & 44.50 & 52.60 & 65.00 & 40.80 & 54.12 \\
FedCoT (r=8,8,8,8,8)  & 68.50 & 45.40 & 53.40 & 65.60 & 41.00 & 54.78 \\
FedCoT (r=16,16,16,16,16) & 69.40 & 45.00 & 54.30 & 65.60 & 41.00 & 55.06 \\
FedCoT (r=32,32,32,32,32) & 66.10 & 45.70 & 54.00 & 66.30 & 41.00 & 54.62 \\
\midrule
FedCoT (r=4,32,32,16,4) & 68.50 & 45.20 & 54.10 & 68.70 & 41.00 & 55.50 \\
\bottomrule
\end{tabular}
\caption{The different performances of \ours{} under different LoRA configurations. ``r" represents the LoRA rank of different clients, corresponding to the clients of the datasets BioASQ, MedMCQA, MedQA, MMLU, and PubMedQA in sequence. The best results are in \textbf{Bold}.}
\label{tab:federated_results}
\end{table*}

%% file: sections/conclusion.tex
\section{Conclusion}

We present FedCoT, a communication-efficient federated framework for reasoning enhancement. FedCoT keeps raw data and actor LLMs local, generates candidate reasoning paths on each client, and trains a lightweight encoder-based discriminator to identify candidates that are more likely to be logically supported. By combining heterogeneous-rank LoRA adaptation and server-side aggregation of lightweight discriminator modules, the framework enables cross-client knowledge sharing without transmitting full actor-model parameters. Empirically, across various benchmarks, \ours{} consistently outperformed federated SFT baselines while reducing communication costs.

%% file: sections/limitations.tex
\section*{Limitations and Future Work}

This study is constrained by available compute and deployment resources: we evaluate \ours{} mainly in cross-silo biomedical QA with moderate client scales and model sizes, rather than in broader or more demanding settings (e.g., large-scale or cross-device deployments). In addition, our discriminator effectively selects high-quality reasoning paths but does not provide a fully granular, process-level assessment; incorporating step-wise evaluation and supervision is left for future work.

\section*{Ethics Statements}
Our framework preserves data locality and shares only aggregated updates. Real-world deployment should incorporate differential privacy and regulatory compliance. This is a research prototype and not a clinical decision tool; any medical application requires rigorous validation and professional oversight. All experiments use publicly available data following their usage agreements, and no patient privacy or personal information is used throughout this process. 


%% file: sections/appendix/algorithm.tex
\section{Algorithm}
\label{ap:algo}

Here, we present the pseudo-code of the entire FedCoT method, see Algorithm~\ref{alg:fedcot_alg}:

\begin{algorithm}[]
\SetAlgoLined
\SetKwProg{Function}{Function}{\string :}{}

\KwIn{
    Total rounds $T$; 
    Local training epochs $E$; 
    Client datasets $\{\mathcal{D}_i\}_{i=1}^N$; 
    Actor LLM $p_\psi$;
    Pretrained discriminator $d_\phi$;
    Number of candidates $K$
}

\caption{\ours{} Algorithm}
\label{alg:fedcot_alg}

\For{$t = 1$ \KwTo $T$}{
    \ForEach{client $i$}{
        Receive last aggregated adapter \(\mathbf W^{t-1}\)
        $(\mathbf{A}_i^t, \mathbf{B}_i^t, \mathbf W^{cls}_i) \leftarrow \textit{LocalUpdates}(i, \mathbf W^{t-1})$\;
    }
    Aggregate LoRA modules and classifier using Equation~\ref{eq:lora_stack},~\ref{eq:aggregate_cls} to get adapter $\mathbf W^t$ \;
}

\Function{\textit{LocalUpdates}($i$, $\mathbf W$)}{
    \ForEach{$(x_j, y_j) \in \mathcal D_i$}{
    Generate reasoning paths using \(p_\psi\):
    $\{(\tau_{j,k}, \hat{y}_{j,k})\}_{k=1}^K$\;
    Construct candidate features $\mathbf h_{j,k}$ and soft labels $s_{j,k}$;
    apply candidate selection to construct the local discriminator-training set\;
}

\For{$e = 1$ \KwTo $E$}{
    Update LoRA modules and classifier using Equation~\ref{eq:seq_cls_loss}\;
}
}

\ForEach{test sample $x_j$}{
    Generate reasoning paths $\{(\tau_{j,k}, \hat{y}_{j,k})\}_{k=1}^K$\;
    Concatenate feature vector $\mathbf{h}_{j,k} = [x_j \parallel \tau_{j,k} \parallel \hat{y}_{j,k}]$\;
    Select answer via Equations~\ref{eq:get_p},~\ref{eq:get_score}, and~\ref{eq:get_answer}\;
}

\KwOut{Final answer \{$\hat{y}_j$\}}
\end{algorithm}

%% file: sections/appendix/length.tex
\section{Analysis of Different Context Length}
\label{ap:length}

\input{tables/truncation}

\input{tables/length_comparison}

To address potential concerns regarding truncation effects under the 512-token constraint (e.g., premature termination before generating complete answer), we extended the maximum generation length to 1024 tokens. This intervention effectively eliminated truncation issues, as evidenced by the near-zero truncation rates in Table~\ref{tab:truncation_rates}. 

Performance comparisons in Table~\ref{tab:length_comparison} reveal consistent accuracy improvements across most configurations after length extension. Our \ours{} method achieved gains of +3.04\% and +1.00\% for \texttt{LLaMA-3-8B-Instruct} and \texttt{Qwen2.5-7B-Instruct}, respectively, while FedIT showed a +0.43\% improvement for \texttt{LLaMA-3-8B-Instruct}. The marginal decline observed for FedIT on \texttt{Qwen2.5-7B-Instruct} (-0.27\%) may be attributable to increased noise in extended reasoning chains.

Critically, \ours{} demonstrates robust effectiveness when controlling for truncation effects, delivering substantial improvements of +9.30\% and +9.16\% over FedIT baselines for \texttt{LLaMA-3-8B-Instruct} and \texttt{Qwen2.5-7B-Instruct}, respectively token length with 1024. These gains exceed those observed under 512-token constraints, validating both the efficacy and robustness of our approach across generation length parameters.

%% file: tables/truncation.tex
\begin{table}[t]
\centering
\begin{tabular}{@{}clc@{}}
\toprule
Max Length & Actor Model & Avg.(\%) \\
\midrule
512 & LLaMA-3-8B-Instruct & 5.93 \\
    & Qwen2.5-7B-Instruct & 26.28 \\
\midrule
1024 & LLaMA-3-8B-Instruct & 0.02 \\
     & Qwen2.5-7B-Instruct & 0.04 \\
\bottomrule
\end{tabular}
\caption{The mean truncation rates on all datasets across maximum generation lengths.}
\label{tab:truncation_rates}
\end{table}

%% file: tables/length_comparison.tex
\begin{table*}[htbp]
\centering
\begin{tabular}{@{}llccccc>{\kern-\tabcolsep}c@{}}
\toprule
Model & Method & BioASQ & MedMCQA & MedQA & MMLU & PubMedQA & Avg.(\%) \\
\midrule
\multirow{4}{*}{\shortstack{LLaMA-3-8B-\\Instruct}}
& FedIT-512 & 42.74 & 47.29 & 53.73 & 71.17 & 13.20 & 45.63 \\
& FedCoT-512 & 65.30 & 45.20 & 56.10 & 54.00 & 41.00 & \underline{52.32} \\
& FedIT-1024 & 45.16 & 46.45 & 54.36 & 70.55 & 13.80 & 46.06 \\
& FedCoT-1024 & 78.20 & 44.40 & 53.70 & 58.30 & 42.20 & \textbf{55.36} \\
\midrule
\multirow{4}{*}{\shortstack{Qwen2.5-7B-\\Instruct}}
& FedIT-512 & 82.26 & 48.48 & 44.30 & 68.71 & 47.20 & 58.19 \\
& FedCoT-512 & 96.80 & 50.00 & 52.50 & 66.30 & 64.80 & \underline{66.08} \\
& FedIT-1024 & 82.26 & 48.55 & 43.28 & 68.10 & 47.40 & 57.92 \\
& FedCoT-1024 & 97.60 & 51.60 & 56.50 & 59.50 & 70.20 & \textbf{67.08} \\
\bottomrule
\end{tabular}
\caption{Performance comparison (\%) across different maximum generation token lengths. The best results are in \textbf{Bold} and the second-highest results are indicated with an \underline{underline}.}
\label{tab:length_comparison}
\end{table*}

%% file: sections/appendix/explore_prm.tex
\section{Discussion on Fine-grained Process-oriented Discrimination}
\label{ap:prm}
\input{tables/steps_info}

We investigate whether process-oriented evaluation could provide more fine-grained signals. Inspired by process reward models~\cite{lightman2023let} in reinforcement learning, we design a step-wise self-evaluation mechanism, where each client model assigns confidence scores to its intermediate reasoning steps.

Surprisingly, as shown in Table \ref{tab:model_step_judgment}, models exhibit strong positivity bias, with self-evaluated positive rate ranging from 81.63\% to 99.98\% across all datasets. This overconfidence persists even in cases where the final answer accuracy is critically low (e.g., only 9.20\% on PubMedQA), indicating that self-assessment fails to distinguish correct from flawed reasoning paths. This may be because models lack reliable internal uncertainty estimation for intermediate steps, and the self-evaluation task, being trained on the same data distribution, inherits the model’s existing biases.




%% file: tables/steps_info.tex
\begin{table}[ht]
\centering
\begin{tabular}{lccc}
\toprule
Dataset    & Positive & Negative  & Ratio(\%)   \\
\midrule
BioASQ     & 5,740 & 145  & 97.54 \\
MedMcQA    & 47,290 & 5,408 & 89.74 \\
MedQA      & 118,719 & 26   & 99.98 \\
MMLU       & 36,676 & 979   & 97.40 \\
PubMedQA   & 3,288  & 740   & 81.63 \\
\bottomrule
\end{tabular}
\caption{Step-wise self-evaluation performance across medical QA benchmarks. Positive: count of reasoning steps judged correct by the model; Negative: count of steps judged incorrect; Ratio: proportion of correct self-evaluation. }
\label{tab:model_step_judgment}
\end{table}

%% file: sections/appendix/open_rm.tex
\section{Performance of General Open-source Reward Model}
\label{ap:open_rm}

\begin{table*}[htbp]
\centering
\begin{tabular}{lccccc}
\toprule
Model & BioASQ & MedMCQA & MedQA & MMLU & PubMedQA \\
\midrule
Baseline & 37.90 & 29.80 & 27.20 & 38.70 & 9.20 \\
Skywork-Reward-V2-Qwen3-0.6B & 31.50 & 24.50 & 20.20 & 32.50 & 4.00 \\
Skywork-Reward-V2-Llama-3.2-1B & 41.90 & 23.90 & 25.80 & 28.80 & 6.40 \\
Skywork-Reward-V2-Qwen3-1.7B & 25.00 & 24.80 & 21.20 & 26.40 & 1.00 \\
Skywork-Reward-V2-Llama-3.1-8B & 16.10 & 12.80 & 7.60 & 16.00 & 0.60 \\
\bottomrule
\end{tabular}
\caption{Performance of open-source SOTA reward models in the medical domain across different benchmarks (in \%).}
\label{tab:medical_reward_models}
\end{table*}

We further verify the applicability of general-domain reward models~\cite{liu2025skywork} in medical task scenarios-specifically focusing on their ability to filter and select high-quality reasoning traces in medical scenarios, and the results are shown in Table~\ref{tab:medical_reward_models}. Although individual gains occur on specific datasets, all evaluated open-source reward models yield lower average accuracy than direct CoT prompting. This reflects the special nature of medical scenarios; crudely and directly utilizing these models cannot transfer their excellent performance from general scenarios.

%% file: sections/appendix/case_study.tex
\section{Case Study}
\label{ap:case}

This case in Table~\ref{tab:case_medqa} involves a 29-year-old man with burning urination (urethritis), acute asymmetric joint pain (right ankle, left knee), bilateral conjunctivitis, and recent antibiotic-treated bloody diarrhea. 

The question asks for the most likely additional finding among four options. The correct answer is B (Tenderness at the insertion of the Achilles tendon), indicative of reactive arthritis triggered by enteric infection (e.g., Shigella/Salmonella).

Generation 1 incorrectly prioritized finger joint pain despite the patient's ankle pain being the critical clue. Reactive arthritis typically involves lower extremities (e.g., Achilles enthesitis), not finger joints. Generation 2 ignored the 2-week latency between diarrhea and joint symptoms—a hallmark of reactive (not septic) arthritis. 

Generation 3 correctly diagnosed reactive arthritis and prioritized Achilles tenderness (B) as the key additional finding, aligning with the patient's ankle pain

Other generation all made wrong answers. 

The discriminator model correctly assigned the highest score (0.834) to generation 3 for its precise pathophysiology: linking Achilles tenderness to reactive arthritis.

Other generations scored 0.62–0.79 for plausible-but-incorrect "abdominal infection" theories, yet the model still ranked them below the correct answer.

\begin{table*}[]
\centering
\vspace{-1ex}
\resizebox{\textwidth}{!}{%
\begin{tabular}{p{20cm}}
\toprule
\textbf{Question}:A previously healthy 29-year-old man comes to the emergency department because of burning with urination for several days. He has also had pain in the right ankle for 3 days and pain and swelling in the left knee for 1 day. Two weeks ago, he had several days of fever and bloody diarrhea, for which he was treated with antibiotics. Examination shows a small left knee effusion and bilateral conjunctival injection. Which of the following is the most likely additional finding in this patient?\textbackslash n A. Palpable mass in the right lower quadrant\textbackslash n B. Tenderness at the insertion of the Achilles tendon\textbackslash n C. Pain on passive extension of the fingers\textbackslash n D. Circular erythematous rash with central clearing
\qquad
\textbf{Ground-Truth Answer}: {\textbf{B}}; \\ \hline

\textbf{Generation 1}: Here is the step-by-step reasoning to answer the question:\textbackslash n\textbackslash nThe patient presents with a combination of symptoms, including burning with urination, pain in the right ankle, and pain and swelling in the left knee. \(\ldots\) The development of new symptoms, such as joint pain and swelling, suggests that the infection may have spread to other parts of the body.\textbackslash n\textbackslash nThe patient's conjunctival injection (inflammation of the conjunctiva) is also a significant finding, as it is a common manifestation of disseminated infection.\textbackslash n\textbackslash nGiven the patient's symptoms and medical history, the most likely diagnosis is reactive arthritis, which is a type of spondyloarthropathy that occurs in response to an underlying infection, often gastrointestinal.\textbackslash n\textbackslash nReactive arthritis typically affects the joints, eyes, and skin, and can cause symptoms such as joint pain, swelling, and inflammation. The patient's bilateral conjunctival injection and left knee effusion are consistent with this diagnosis.\textbackslash n\textbackslash nThe most likely additional finding in this patient would be:\textbackslash n\textbackslash n\#\#\#\# C. Pain on passive extension of the fingers\textbackslash n\textbackslash nThis symptom is a common manifestation of reactive arthritis, as the condition can cause inflammation and stiffness in the hands and fingers. Passive extension of the fingers would likely cause pain and stiffness, which would be consistent with the patient's underlying diagnosis.\textbackslash n\textbackslash nThe other options are not as likely:\textbackslash n\textbackslash nB. Tenderness at the insertion of the Achilles tendon: Achilles tendonitis is a possible diagnosis, but it is not the most likely explanation for the patient's symptoms given the patient's medical history and conjunctival injection.\textbackslash n\textbackslash n\\
\textbf{Prediction 1}: \textbf{C}; \quad  \quad \textbf{Score 1:} 0.77344\\ \hline 

\textbf{Generation 2}: Here is the step-by-step reasoning to answer the question:\textbackslash n\textbackslash nThe patient presents with a complex set of symptoms, including burning with urination, pain in the right ankle, and pain and swelling in the left knee. The patient also reports a history of fever and bloody diarrhea two weeks ago, which was treated with antibiotics.\textbackslash n\textbackslash n\(\ldots\)Step 4: Consider the possible causes of the patient's symptoms in the context of the possible sources of infection.\textbackslash n\textbackslash nGiven the patient's history of bloody diarrhea and the presence of symptoms in the joints and eyes, it is possible that the patient has developed a septic arthritis, which is an infection of the joint that is caused by bacteria that have spread from another part of the body.\textbackslash n\textbackslash nStep 5: Evaluate the patient's symptoms in relation to the possible causes of septic arthritis.\textbackslash n\textbackslash nThe patient's symptoms of pain and swelling in the joints, as well as the presence of a small left knee effusion, are consistent with septic arthritis. The patient's bilateral conjunctival injection also suggests that the infection may have spread to the eyes.\textbackslash n\textbackslash nStep 6: Consider the most likely additional finding in the patient.\textbackslash n\textbackslash nGiven the patient's symptoms and the possible causes of septic arthritis, the most likely additional finding is a palpable mass in the right lower quadrant, which could indicate an abscess or a collection of pus in the right lower quadrant of the abdomen.\textbackslash n\textbackslash n\#\#\#\# A. Palpable mass in the right lower quadrant \\
\textbf{Prediction 2}: \textbf{A}; \quad  \quad \textbf{Score 2:} 0.717285\\ \hline 

\textbf{Generation 3}: Here is the step-by-step reasoning:\textbackslash n\textbackslash nThe patient presents with a combination of symptoms, including burning with urination, pain in the right ankle, and pain and swelling in the left knee. This suggests that the patient may have a systemic infection that is affecting multiple parts of the body.\textbackslash n\textbackslash nThe patient had a recent episode of bloody diarrhea, which is a common symptom of gastrointestinal infections. The fact that he was treated with antibiotics suggests that the infection was likely bacterial in nature.\textbackslash n\textbackslash nThe patient's symptoms are consistent with a urinary tract infection (UTI), as burning with urination is a common symptom of UTIs. The presence of a small left knee effusion and bilateral conjunctival injection suggests that the infection may have spread to the joints and eyes, respectively.\textbackslash n\textbackslash nThe patient's symptoms are also consistent with reactive arthritis, a condition that can occur as a complication of certain bacterial infections, such as Salmonella or Shigella. Reactive arthritis is characterized by joint pain and swelling, as well as conjunctivitis and urethritis.\textbackslash n\textbackslash nGiven the patient's symptoms and medical history, the most likely additional finding is:\textbackslash n\textbackslash nB. Tenderness at the insertion of the Achilles tendon\textbackslash n\textbackslash nThis is because reactive arthritis often affects the joints, including the ankles, and the Achilles tendon is a common site of tenderness in this condition.\textbackslash n\textbackslash n\#\#\#\# B. Tenderness at the insertion of the Achilles tendon \\
\textbf{Prediction 3}: \textbf{B}; \quad  \quad \textbf{Score 3:} 0.833984\\ \hline

\textbf{Generation 4}: ...... \\
\textbf{Prediction 4}: \textbf{A}; \quad  \quad \textbf{Score 4:} 0.621094\\ \hline

\textbf{Generation 5}: ...... \\
\textbf{Prediction 5}: \textbf{A}; \quad  \quad \textbf{Score 5:} 0.785645\\ \hline

\textbf{Generation 6}: ...... \\
\textbf{Prediction 6}: \textbf{A}; \quad  \quad \textbf{Score 6:} 0.781250\\ \hline

\textbf{Generation 7}: ...... \\
\textbf{Prediction 7}: \textbf{A}; \quad  \quad \textbf{Score 7:} 0.777344\\ \hline

\textbf{Generation 8}: ...... \\
\textbf{Prediction 8}: \textbf{C}; \quad  \quad \textbf{Score 8:} 0.717285\\

\bottomrule
\end{tabular}%
}
\caption{A case study on MedQA dataset.}
\label{tab:case_medqa}
\end{table*}

%% file: sections/appendix/model_improvement_potential.tex
\section{Improvement Potential}\label{ap:improv}

\begin{table*}[!htbp]
\centering

\setlength{\tabcolsep}{4pt} 
\renewcommand{\arraystretch}{1.3} 
\small 
\begin{tabular}{l*{12}{c}c}
\toprule
\multirow{2}{*}{\textbf{Model}} 
& \multicolumn{2}{c}{\textbf{bioasq}} 
& \multicolumn{2}{c}{\textbf{medmcqa}} 
& \multicolumn{2}{c}{\textbf{medqa}} 
& \multicolumn{2}{c}{\textbf{mmlu}} 
& \multicolumn{2}{c}{\textbf{pubmedqa}} 
& \multicolumn{2}{c}{\textbf{average}} 
& \multirow{2}{*}{\textbf{$\Delta$}} \\
\cmidrule(lr){2-3} \cmidrule(lr){4-5} \cmidrule(lr){6-7} 
\cmidrule(lr){8-9} \cmidrule(lr){10-11} \cmidrule(lr){12-13}
& \textbf{p@1} & \textbf{p@k} 
& \textbf{p@1} & \textbf{p@k} 
& \textbf{p@1} & \textbf{p@k} 
& \textbf{p@1} & \textbf{p@k} 
& \textbf{p@1} & \textbf{p@k} 
& \textbf{p@1} & \textbf{p@k} & \\
\midrule

LLaMA-3-8B    & 37.90 & 70.97 & 29.84 & 68.18 & 27.18 & 70.46 & 38.65 & 71.78 &  9.20 & 42.00 & 28.55 & 64.68 & 36.12 \\
LLaMA-3.1-8B  & 25.81 & 64.52 & 35.07 & 72.22 & 32.05 & 73.53 & 39.26 & 77.30 & 15.20 & 54.80 & 29.48 & 68.47 & 38.99 \\
Qwen2.5-7B      & 73.39 & 98.39 & 43.75 & 70.98 & 29.46 & 66.54 & 50.31 & 78.53 & 38.80 & 73.40 & 47.14 & 77.57 & 30.43 \\
\midrule\midrule
Qwen2.5-3B      & 10.48 & 46.77 & 18.98 & 55.41 &  5.03 & 23.80 & 26.38 & 65.64 &  1.20 &  7.40 & 12.41 & 39.81 & 27.39 \\
LLaMA3.2-3B  & 31.45 & 70.16 & 31.41 & 65.57 & 20.03 & 59.62 & 32.52 & 70.55 &  4.80 & 24.20 & 24.04 & 58.02 & 33.98 \\
\midrule\midrule
Qwen-1.5B    &  0.81 &  9.68 &  1.65 & 11.09 &  1.73 & 15.48 &  0.61 & 10.43 &  0.00 &  0.40 &  0.96 &  9.41 &  8.46 \\
LLaMA3.2-1B  &  0.00 &  0.00 &  0.10 &  0.36 &  0.00 &  0.00 &  0.00 &  0.00 &  0.00 &  0.00 &  0.02 &  0.07 &  0.05 \\
Qwen2.5-0.5B    &  0.00 &	0.81 &	1.63 &	8.92 &	0.47 &	5.89 &	1.23 &	7.98 &	0.60 &	3.20 &	0.78 &	5.36 &	4.57 \\

\bottomrule
\end{tabular}
\caption{Model performance (in \%) on biomedical test sets demonstrating accuracy potential through multiple sampling. 
The table shows pass rates at first sample (p@1) and after $k$ samples (p@k), with $\Delta = \text{p@k} - \text{p@1}$ 
indicating accuracy improvement potential. }
\label{tab:model_performance}
\end{table*}

Table~\ref{tab:model_performance} shows the improvement of the test set of multiple models with three different parameter levels under multiple sampling, where the models with less than 3B parameters can hardly achieve performance improvement through multi-sampling. 
The model with 3B parameters shows a significantly higher improvement under multiple samplings than models with less than 3B parameters, and its improvement performance is relatively close compared to that of the model with 7B parameters.

In addition, we can see that in the case of less than 3B, the Qwen series of models can still achieve a slight improvement, while the LLaMA series of models basically have no improvement at all. This may be due to different treatments during pre-training.

Overall, whether for the Qwen series or the LLaMA series, the larger the model parameter count, the better the basic performance under the corresponding dataset conditions, and the greater the improvement obtained through sampling. This reflects that the basic capabilities of the model are a very crucial part for further subsequent improvement. If the basic capabilities of the base model are too low, then even more sampling cannot achieve significant improvement.

%% file: sections/appendix/detailed_datasets_info.tex
\section{Datasets Information}
\label{ap:detailed_datasets_info}

We will showcase the medical datasets used for generating CoT candidates during the training phase and the medical datasets used for evaluation during the testing phase here.

\begin{itemize}
    \item PubMedQA~\cite{jin2019pubmedqa}: A biomedical question-answering dataset based on PubMed abstracts, containing 1,000 expert-annotated multiple-choice (yes/no/maybe) biomedical question-answering samples. Its knowledge source involves 211,300 PubMed papers. This task requires judging the answer to the research question based on the abstract content. The average lengths of the question stem and context are 14.4 and 238.9 tokens respectively. It is distributed under MIT license.
    \item BioASQ~\cite{tsatsaronis2015overview}: A large-scale biomedical semantic indexing and question-answering dataset, including two tasks: information retrieval (Task A) and machine reading comprehension (Task B). This paper focuses on 618 binary judgment (yes/no) reading comprehension questions in Task B in the recent five years (2019 - 2023). The average length of the question tokens is 17. Similar to PubMedQA, this dataset supports machine reading comprehension research by providing standard answer text segments. It is distributed under the CC-BY-4.0 license.
    \item MMLU-Med~\cite{hendrycks2020measuring}: A medical reasoning specialized dataset extracted from the multi-task language understanding benchmark MMLU (hereinafter referred to as MMLU for short). In this paper's experiments, seven medical-related fields are selected: clinical knowledge, college biology, college medicine, high school biology, medical genetics, professional medicine, and virology. The reasoning ability of the model in professional medical scenarios is mainly investigated. It is distributed under MIT license.
    \item MedMCQA~\cite{pal2022medmcqa}: A large-scale multiple-choice medical question-answering dataset that integrates real question resources from the All India Institute of Medical Sciences (AIIMS) and the National Eligibility cum Entrance Test for Postgraduate (NEET-PG) in India. This dataset contains over 194,000 high-quality medical questions, covering 2,400 health topics and 21 medical discipline areas, featuring significant topic diversity. The average token length of the questions is 12.77. It is distributed under Apache-2.0 license.
    \item MedQA~\cite{jin2021disease}: A clinical medicine question bank based on the United States Medical Licensing Examination (USMLE), containing 12,723 questions sourced from 18 widely used authoritative clinical medicine textbooks. The questions cover a rich variety of clinical medicine topics and require professional-level reasoning by integrating multi-source evidence. The average length of the questions is 116.6 tokens, and the average length of the options is 3.5 tokens. It is distributed under the CC-BY-4.0 license.
    \item GPQA (Medical)~\cite{rein2024gpqa}: As the biomedical domain subset derived from the Graduate-Level QA dataset, GPQA (Medical) is characterized by expert-curated multiple-choice questions of high difficulty. These questions are specifically designed to necessitate advanced reasoning capabilities, with an intentional design objective to avoid being answerable through simple search-engine queries—thereby eliminating over-reliance on surface-level information retrieval and emphasizing the need for in-depth domain understanding. 
    \item MedBullets\cite{chen2024benchmarking}: This dataset comprises practice questions curated from the MedBullets medical education platform, a specialized resource for medical learning and skill assessment. Owing to their alignment with high-difficulty medical knowledge application scenarios, these subsets function as rigorous evaluation benchmarks for assessing the performance of models in the biomedical domain.
    \item MedXpertQA~\cite{zuo2025medxpertqa}: This dataset consists of 50 medical reasoning questions, all of which are authored by domain experts and structured to require multi-step analytical processes. The expert-authored nature of the questions ensures high validity and authenticity of the scenarios, while the multi-step reasoning framework aligns with the complexity of real-world clinical decision-making, thereby enabling rigorous, nuanced assessments of model performance in biomedical reasoning tasks. It is distributed under MIT license.
\end{itemize}

%% file: sections/appendix/prompt_template.tex
\section{Prompt Template}
\label{ap:prompts}

We present the one-shot CoT prompt for generating candidate reasoning paths in Figure~\ref{prompt_template}. This carefully designed template guides the model in rigorous reasoning and follows required formatting for easy answer extraction and evaluation. 
We also presented the prompt template used by LLM-as-Judge to evaluate the factual fidelity, explainability, and overall quality of reasoning paths, as shown in Figure \ref{prompt_template2}, with evaluation scores ranging from 1 to 10.

\begin{figure}[!htbp]
\begin{minipage}{\linewidth}
\begin{framed}
\noindent\textbf{The Prompt Template for one-shot CoT}

\medskip
\noindent To answer the following question, provide a highly detailed and comprehensive step-by-step reasoning. \\
Your explanation should break down the problem into clear, logical steps, with each step on a new line.\\
The goal is to construct a robust and self-contained argument that fully justifies the final conclusion.\\
After the reasoning, present the final answer. Please model the overall structure on the provided example.

\medskip
\noindent\textbf{Example:}\\
\textbf{Q:} \\
Primary Colonization of plaque is dominated by \\
A. Facultative Aerobic Gram+ve rods\\
B. Facultative Anaerobic Gram-ve rods\\
C. Facultative Aerobic Gram-ve cocci\\
D. Facultative Anaerobic Gram+ve cocci\\

\noindent\textbf{A:} \\
The oral cavity is a complex ecosystem where various microorganisms coexist, and plaque is a biofilm that forms on tooth surfaces.\\
Primary colonization of plaque is dominated by Streptococcus species, which are Gram-positive, facultative anaerobic cocci.\\
These Streptococcus species are able to adhere to the tooth surface and multiply, creating a matrix for other microorganisms to colonize.\\
\#\#\#\# D. Facultative Anaerobic Gram+ve cocci

\medskip
\noindent\textbf{Here is the question:}\\
\textbf{Q:}\\
\texttt{\{question\}}\\
\texttt{\{options\}}
\end{framed}

\captionsetup{width=\linewidth}
\caption{Template for Multiple-choice Questions (MCQs). 
Placeholders \texttt{question} and \texttt{options} denote the MCQ stem and options.}
\label{prompt_template}
\end{minipage}
\end{figure}

\begin{figure}[!htbp]
\centering
\begin{minipage}{\linewidth}
\begin{framed}
\noindent\textbf{The Prompt Template for 3 dimensional judgments.}

\medskip
\noindent 

You are a medical expert evaluating AI responses to medical questions. Please evaluate the following response based on three criteria and provide scores from 1-10 (10 being best). \\
Medical Question: \texttt{\{question\}} \\
AI Response: \texttt{\{response\}} \\

Evaluation Criteria: \\
1. FACTUAL FAITHFULNESS (1-10): Assess whether the medical information is accurate and consistent with established medical knowledge. Penalize hallucinations, unsupported claims, or factual inaccuracies. \\
2. EXPLANATORY QUALITY (1-10): Evaluate how well the response explains medical concepts, provides reasoning, and makes complex information understandable. Consider clarity and educational value. \\
3. OVERALL QUALITY (1-10): Judge the comprehensive quality including completeness, relevance to the question, professionalism, and practical usefulness for medical understanding. \\

Output Format - Provide your evaluation in this exact JSON format: \\
\{\{ \\
    `factual\_faithfulness': \{\{ \\
        `score': 7, \\
        `reasoning': `Brief explanation of strengths and weaknesses in factual accuracy'
    \}\}, \\
    `explanatory\_quality': \{\{ \\
        `score': 8, \\
        `reasoning': `Brief analysis of explanatory effectiveness' \\
    \}\}, \\
    `overall\_quality': \{\{ \\
        `score': 7, \\
        `reasoning': `Overall assessment of the response's medical value' \\
    \}\} \\
\}\} \\
Ensure scores are integers from 1-10. Be strict but fair in medical accuracy assessment.
\end{framed}

\caption{Template for 3D judgments. 
Placeholders \texttt{question} and \texttt{response} denote the medical question and response.}
\label{prompt_template2}
\end{minipage}
\end{figure}